\documentclass[letterpaper, 10 pt, conference]{ieeeconf}  % Comment this line out if you need a4paper

\IEEEoverridecommandlockouts                              % This command is only needed if 
\usepackage{times}

\usepackage{tabularray}
\usepackage{multicol}
\usepackage{hyperref}
\usepackage{color}

\usepackage{graphicx}

 \usepackage{algorithm}
\usepackage{algorithmicx}
\usepackage[noend]{algpseudocode}
\usepackage{wasysym}
\usepackage{diagbox}
\usepackage{multirow}
\usepackage{caption}
\usepackage{subfigure}
\newcommand{\RNum}[1]{\uppercase\expandafter{\romannumeral #1\relax}}
\usepackage{rotating}
\usepackage[normalem]{ulem}

\title{\LARGE \bf
How Much Progress Did I Make? An Unexplored Human Feedback Signal for Teaching Robots
}

\author{Hang Yu$^{1}$, Qidi Fang$^{1}$, Shijie Fang$^{1}$,  Reuben M. Aronson$^{1}$, and Elaine Schaertl Short$^{1}$ % 
\thanks{
{$^{1}$ Tufts University, Medford, MA 02144, USA}}%
}

\begin{document}

\maketitle
\thispagestyle{empty}
\pagestyle{empty}

%%%%%%%%%%%%%%%%%%%%%%%%%%%%%%%%%%%%%%%%%%%%%%%%%%%%%%%%%%%%%%%%%%%%%%%%%%%%%%%%
\begin{abstract}
Enhancing the expressiveness of human teaching is vital for both improving robots' learning from humans and the human-teaching-robot experience. 
In this work, we characterize and test a little-used teaching signal: \textit{progress},  designed to represent the completion percentage of a task. 
We conducted two online studies with 76 crowd-sourced participants and one public space study with 40 non-expert participants to validate the capability of this progress signal. 
We find that progress indicates whether the task is successfully performed, reflects the degree of task completion, identifies unproductive but harmless behaviors, and is likely to be more consistent across participants.  Furthermore, our results show that giving progress does not require extra workload and time. 
An additional contribution of our work is a dataset of 40 non-expert demonstrations from the public space study through an ice cream topping-adding task, which we observe to be multi-policy and sub-optimal, with sub-optimality not only from teleoperation errors but also from exploratory actions and attempts.  The dataset is available at \url{https://github.com/TeachingwithProgress/Non-Expert\_Demonstrations}.

\end{abstract}

%%%%%%%%%%%%%%%%%%%%%%%%%%%%%%%%%%%%%%%%%%%%%%%%%%%%%%%%%%%%%%%%%%%%%%%%%%%%%%%%
\section{Introduction}

Robots have already firmly become part of our daily lives, making it crucial to learn from users, especially non-expert users. 
Learning from Demonstration (LfD) enables robots to learn new skills by observing expert policies \cite{ brown2020better, chen2021learning} while 
Learning from Human Feedback (LfHF) allows robots to adapt to human preferences or correct wrong behaviors by learning or shaping a policy \cite{cui2021empathic, cederborg2015policy, knox2009interactively}.
More recent work has further shown that using human feedback and demonstrations together can make learning even more effective by reducing the data needs for human feedback \cite{ibarz2018reward} and loosening the requirements of demonstrations to be near-optimal \cite{brown2019ranking}. 
However, while interest in learning fully or partially from humans is high, there is relatively little research on what the most 
%the expected quality of demonstrations in previous work is still high, and 
effective forms of human feedback are, especially in \emph{combination} with human demonstrations.

Human feedback and human demonstrations can be complementary due to the difference in human knowledge they carry. 
Demonstrations carry relatively dense and global information including policies and goals, and tend to be less accurate \cite{palan2019learning}. Human feedback carries relatively sparse and local information such as the correctness or a rating of a robot's action, and
giving high-quality feedback can be much easier than giving high-quality demonstrations \cite{arzate2020survey}.
Perfect demonstrations are hard to obtain while purely learning from human feedback requires many human labels, often obtained at significant time and expense.
To address these challenges and improve the quality of learning, human demonstrations can be combined with human feedback: 
demonstrations can be used to train an initial policy to improve the sample efficiency of feedback \cite{ibarz2018reward}, and feedback can be used to refine the policy learned from demonstrations \cite{palan2019learning}.

In previous work using human feedback alongside demonstrations, the forms of human feedback used are directly adapted from LfHF \cite{ibarz2018reward, palan2019learning}, which might not be effective for evaluating a demonstration. 
The quality of a demonstration or a partial demonstration is typically assessed by comparing it to another \cite{palan2019learning, brown2019ranking, brown2020better}. 
This approach of comparing or ranking demonstrations is extremely hard for naive users, especially when the trajectory is only a partial demonstration \cite{laidlaw2021uncertain}. 
It also neglects the objective quality of a demonstration or a partial demonstration itself: a pair of demonstrations might both be good or both bad, making preferences difficult to provide. This is especially true for non-expert demonstrations: non-expert demonstrations can be noisy, multi-policy, and yet still succeed (we show this in \autoref{sectionVC}).
While feedback like binary evaluation and scalar feedback is capable of assessing the quality of a demonstration, 
comparative information is unavailable for binary feedback and is unreliable for scalar feedback \cite{yu2023thumbs}.

\begin{figure}
    \centering
    \includegraphics[width = 0.47\textwidth]{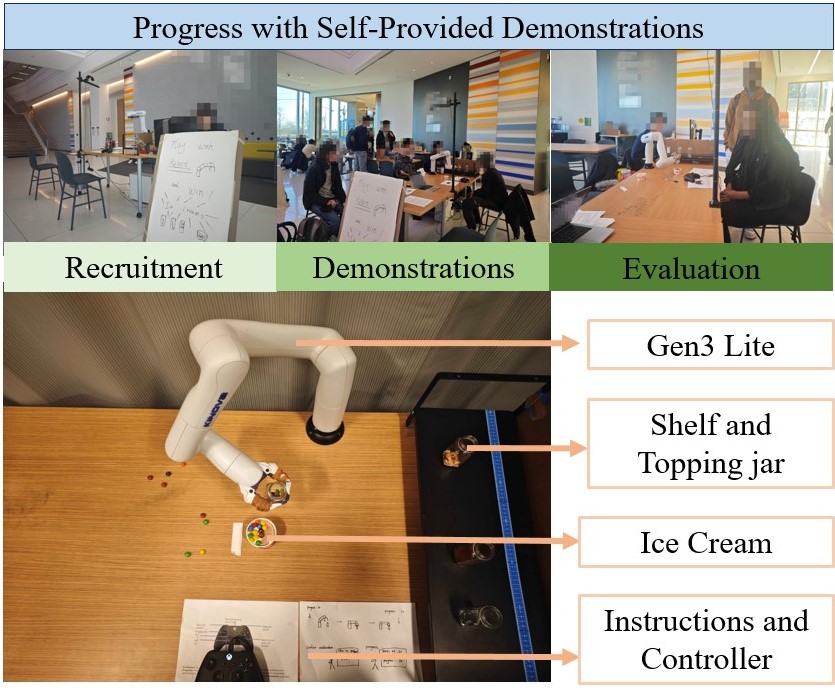}
    \caption{Public space study with an ice cream topping-adding task to collect demonstrations and progress from non-experts.}
    \label{page1}
\end{figure}

In this work, we characterize a novel type of human feedback for robot learning: \textit{progress}, which is used to capture the completion of a task. 
%and can be used alone as a reward signal or to augment the information from demonstrations. 
We show that progress can indicate the extent of task completion, determine if a task is completed, and be robust to unproductive behaviors. Furthermore, when compared to scalar feedback, progress is more consistent when demonstrations are noisy and does not need extra workload and time. 
We show the capability of progress through two online studies and one in-person study.
For the online studies, we recruit 76 crowd-sourced workers to provide scalar feedback and progress 
with pre-recorded expert demonstrations over three simple tasks and one long-horizon task. 
For the in-person study, we recruit 40 passersby to provide demonstrations in an ice cream topping-adding task, shown in \autoref{page1}. The demonstrators then provide progress and scalar feedback to their own demonstrations.

The main contribution of this work is to demonstrate that \textit{progress} has information beyond rating and ranking, and has great potential in interactive learning.
Our validation studies covered a wide range of scenarios and involved 116 participants in total.
Our results also showed that non-expert demonstrations are multi-policy and sub-optimal, but sub-optimal in a meaningful way.
Finally, we released a dataset online with 40 non-expert labeled demonstrations from a public space study, which may better-reflect the types of demonstrations that can be expected from real-world deployments than typical expert or in-lab demonstration datasets. 

\section{Background}

By leveraging human knowledge, interactive machine learning allows learning agents to adapt to the needs of individual users and improve sample efficiency \cite{arzate2020survey}. 
Human knowledge can take a variety of forms, such as semantic representation \cite{kostavelis2013learning}, numerical feedback \cite{knox2009interactively}, eye gaze \cite{saran2020understanding}, gestures \cite{yanik2013gesture}, facial expressions \cite{cui2021empathic}, and demonstrations \cite{ravichandar2020recent}. 
In this work, we focus on human feedback and human demonstrations. Separately, each of these approaches has significant limitations: on the one hand, inferring a policy from human feedback requires a large number of interactions \cite{knox2009interactively, arakawa2018dqn, warnell2018deep}, and on the other, error-free demonstrations are rare and expensive to obtain in the real world \cite{ravichandar2020recent}.
To address this, many approaches in Interactive Machine Learning seek to combine human feedback and human demonstration to compensate for the limitations of each.
% There is a common assumption in many prior works that expert demonstrations are optimal \cite{ning2022learning}. 
% 

% One approach to mitigate the imperfection is to use extra information to distinguish the quality difference between demonstrations by either using generative methods \cite{brown2020better, wang2021learning} or asking humans. 

% 
% However, 
%
% Previous work has shown that consolidating human demonstration with human feedback, such as human ranking \cite{brown2019ranking}, human corrections \cite{iturrate2017learning}, weight labels \cite{ning2022learning, wang2021learning}, and human preferences \cite{biyik2022learning, palan2019learning}, can facilitate effectively learning a stable policy or accurate reward function. 

\noindent \textbf{Learning from Human Feedback}
  Learning from Human Feedback has emerged as a promising technology for robots or machine-learning agents to learn from humans via interactions \cite{casper2023open, yu2021active} . 
 LfHF, in general, refers to methods that have three components: feedback collection,  policy or reward shaping, and policy optimization.  
 Human feedback can be in a variety of forms, such as verbal \cite{goyal2019using, kuhlmann2004guiding}, numerical \cite{cederborg2015policy, knox2009interactively, arumugam2019deep}, and implicit \cite{cui2021empathic, xu2021accelerating, TAN2020113043}.
 % In this work, we mainly focus on numerical explicit human feedback, since these signals are most commonly used in combination with human demonstrations. 
Three representative works of learning from explicit human feedback are  Policy Shaping \cite{cederborg2015policy}, TAMER \cite{knox2009interactively}, and  Preference-based policy learning \cite{akrour2011preference}.
% Policy Shaping \cite{cederborg2015policy} shapes the learned policy by querying users to indicate whether the robot's action was correct. TAMER \cite{knox2009interactively} replaces the environment rewards with human evaluations.
% Preference-based policy learning learns a policy based on preferences over two candidate policies.
Human feedback has also been applied to modern Large Language Models (LLMs) \cite{2023arXiv230308774O} to further improve the performance of trained models by having humans rate the outputs with binary critiques. 
Despite a large body of work that has been done, human feedback is mostly used as reward signals. 
% Thus, our work differs from prior work by proposing a new form of human feedback that carries not only rewards but also descriptions of task completion. 

\noindent \textbf{Learning from Human Demonstrations} 
In robotics, Learning from Demonstrations (LfD) is a method that facilitates robots to learn new skills by imitating humans \cite{chernova2014robot}.
The use of LfD offers several advantages, including eliminating the need for expert programming \cite{zhu2018robot}, high data efficiency \cite{ravichandar2019skill}, safety for learning \cite{umlauft2017learning}, and guaranteed performance \cite{ijspeert2013dynamical}.
The research interest in teaching robots via demonstrations has steadily advanced. 
% Even for complex continuous control tasks, 
LfD methods are capable of producing optimal behaviors with clean demonstrations and sufficient error-free demonstrations \cite{sasaki2018sample}. 
However, due to the optimal assumption on the demonstrations, LfD methods like generative adversarial imitation learning \cite{ho2016generative} or behavior cloning \cite{fang2019survey} failed to acquire optimal policy for many robot tasks since even human experts would make mistakes while providing demonstrations \cite{wu2019imitation}. 
% Thus, learning from non-optimal demonstrations has become popular. 
% One straightforward solution is screening out non-optimal demonstrations, but screening often involves a lot of human effort and significantly reduces sample efficiency because optimal demonstrations are rare \cite{sasaki2020behavioral}.
Weighting \cite{sasaki2020behavioral, wu2019imitation,chen2021learning} or ranking \cite{ brown2020better, wang2021learning} demonstrations are considered to be robust methods of learning from noisy demonstrations. 
However, learning from multiple users and learning from imperfect demonstrations are still challenging problems \cite{ravichandar2020recent}. One previous work used a technique they refer to as ``reward sketching''; in practice the annotators were instructed to provide progress \cite{cabi2019scaling}.  While their work demonstrates the potential of progress for guiding learning, it did not closely investigate the properties of progress or take full advantage of it as a teaching signal, instead using large numbers of these ``reward sketching'' annotations as a loose approximation for a dense reward function.

\noindent \textbf{Using Human Demonstrations and Human Feedback}
Recent work has demonstrated that combining human feedback and human demonstrations could overcome many disadvantages of using one of them solely, including safety \cite{brown2020safe}, sample efficiency \cite{ibarz2018reward}, and accuracy \cite{palan2019learning}. 
Specifically, work from \cite{ibarz2018reward} used demonstrations to train an initial model for efficiently collecting preferences from users. Work from \cite{palan2019learning} built on \cite{ibarz2018reward} and used a model-based method to reduce data from humans. 
Using human rankings, work from \cite{brown2019ranking} has achieved super-human demonstration performance.
Although prior work has achieved great success in consolidating human feedback with human demonstrations, the source of human feedback and human demonstrations are likely from experts or pre-trained agents \cite{huang2024modeling}, and 
the human feedback they used could be more informative.
Our work differs from prior work by focusing on non-experts.
We conducted our studies with crowd-sourced workers and random passersby. 
We showed that progress
is informative and consistent when human demonstrations are multi-policy and non-optimal.

\section{Progress}
Our goal is to improve learning from human feedback with a new teaching signal: progress. 
% As previously stated, 
% humans the feedback needs to be more informative.
% We propose` a novel teaching signal \textit{progress} that can be collected along with demonstrations and carries more information. 
In this section, we first define progress, and then introduce our hypothesis. 

\subsection{Progress}
We hypothesize that \textit{progress} provides complementary information to demonstrations beyond rewards. We characterize \textit{progress} as the accumulative task completion rate over a task based on the current observation, ranging from totally incomplete to complete fully. 
Our intuition is that:
    \textit{A teaching signal would be more robust to sub-optimal demonstrations and more consistent among users if human teachers could have objective references while providing the teaching signal. }
For progress specifically, 
users can use start states and finish states as references. 
% which makes it more consistent than scalar feedback or preference which are based on users' own intuition.
In this work, 
 we use the \textit{progress} signal as a range from 0 to 100. 
 A \textit{progress} value of 0 indicates that the task has not yet begun, while a value of 100 signifies task completion. In between, we expect that for any given task $t$, current state $s$, any action $a_i$ and $a_j$, and any previous state $s_i$ and $s_j$:
\begin{equation}
    prog_t(s_i,a_i,s) = prog_t(s_j, a_j, s) \forall i,j
\end{equation}
Ideally, progress is independent of the path taken to the state, irrespective of the sequence of preceding states.
However, human feedback is known to be noisy and can only be considered consistent if we view it as an approximate value \cite{yu2023thumbs}.
Thus, we collect progress by presenting users 
 with a trajectory instead of a single state to increase reliability.  
% Thus, in this work, we refer to the progress we collected as $prog^*_p(s)$ in which $prog^*_p(s)$ is different from person to person and will be affected by human factors. For any given state $s, s' \in S$ We assume that:
% \begin{equation}
%     prog^*_p(s) > prog^*_p(s')\text{, if } prog(s) > prog(s')
% \end{equation}

\subsection{Hypotheses}
We expect that people naturally estimate task completion in their daily lives, so progress should be not hard to give.
We also expect that participants would use progress to describe the completion degree of a task.
We hypothesize:
% the increase and decrease in progress would indicate if the robot's behavior is leading the task toward completion or away from completion. 
% As a result, progress provided at the end of a trajectory should reflect if the task is completed. 
% Moreover, we expect the
% the assessment of progress will partially/fully be based on an objective standard: task completion, resulting in that progress being more consistent among users than rating-based measurement  
% We expect that giving
% progress has specific start states and finish states as references, 
% which makes it more consistent than scalar feedback or preference which are based on users' own intuition.

\noindent \textbf{H1.} Giving progress does not require extra workload and extra time compared to giving scalar feedback.

\noindent \textbf{H2.} Progress describes the completion rate of a task. 

\noindent \textbf{H3.} Progress could correctly indicate if the task is complete even if the robot has made mistakes. 

% \noindent \textbf{H4.} Progress is robust to unproductive behavior, backtracking, or mistakes. 

\noindent \textbf{H4.} Progress is more consistent than scalar feedback when demonstrations are non-optimal.

\section{Progress with non-self-provided demonstrations}
We first crowd-sourced users to provide progress with pre-recorded demonstrations for our online studies. 
This allows us to explore the effectiveness of utilizing progress alone, 
and examine its applicability across a range of intricate scenarios.
% Some of the scenarios were later found to be prevalent among participants in our public space.
% The scenario we used for our online study hawhich we found common in our public space study.  
We conducted two online studies:
Online study \RNum{1}  uses three simple tasks to evaluate the workload of giving progress, while
online study \RNum{2} involves a long-horizon task comprising six sub-tasks and five scenarios to assess the utility of progress.  

% Thus, before using the information from progress for learning, we conducted a two-phase online study to investigate the workload of giving progress and the capability of progress from naive users. 

\subsection{Online study setups}
First, we validated the workload of providing progress and verified that progress contains unique information relative to scalar feedback in a range of 0 to 100.  
% Humans, especially non-experts, might find giving reasonable progress challenging resulting in requiring significantly more effort from users or little meaningful information in progress.
For each study, we recruited two groups of participants from an online platform.
One group of participants was only asked to give progress and the other group of participants was only asked to give scalar feedback. 
Participants in two groups watched the same demonstrations in the same order. 
\begin{figure}
    \centering
    \includegraphics[width = 0.46\textwidth]{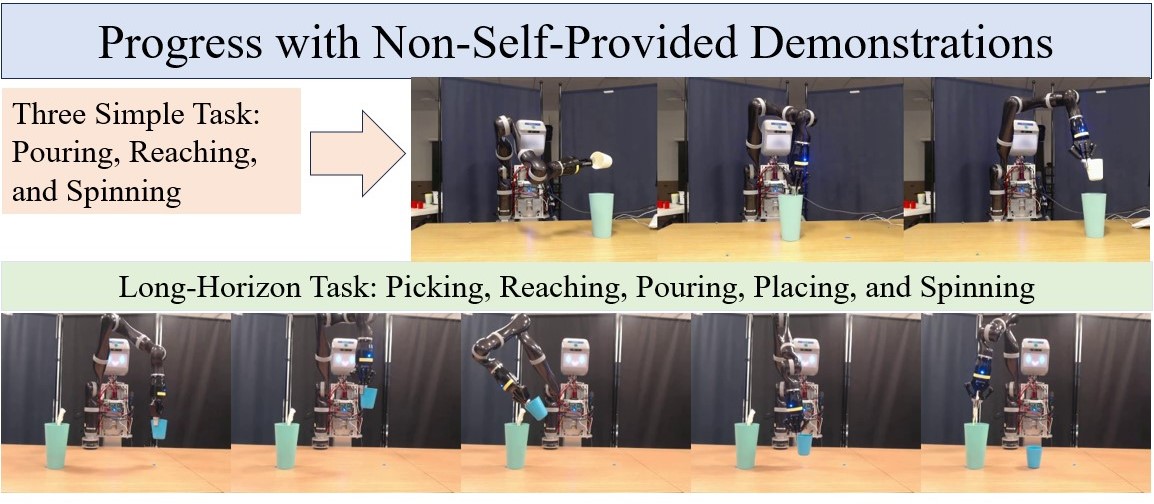}
    \caption{Online Study Setups. Three simple tasks for comparing the workload of giving progress, and a long-horizon task for comparing the applicability of progress.}
    \label{fig:online}
\end{figure}

\subsubsection{Online study \RNum{1}}
We recruited two groups of 20 participants from Amazon Mechanical Turk to provide progress annotations to a robot performing three related tasks. 
The three tasks we used are reaching, pouring, and spinning, shown in \autoref{fig:online}, which are sub-tasks of the task we used in the online study \RNum{2}.
For each task, participants gave 10 progress or 10 scalar feedback annotations. 
After giving all progress or scalar feedback, a NASA TLX \cite{hart2006nasa} questionnaire was given to measure the workload. 
% In phase one, we only used simple tasks to reduce the cognitive load from the tasks.
All demonstrations are perfect demonstrations except the robot made one mistake during the spinning task at step 5. 
\subsubsection{Online study \RNum{2}} 
In study \RNum{2}, we used a long-horizon task with five representative types of variation in task performance to show the capability of progress. 
The task we use is a tea steeping task (also shown in \autoref{fig:online}), which is a task combining picking, reaching, pouring, and spinning.
We recruited another two groups of 18 participants from Prolific, a crowdsourcing platform for scientific research, to give scalar feedback or progress. 
% We switched to another platform to reduce the overlap in the participant pool. 
We chose five demonstrations each representing a distinct scenario:
Perfect (everything is performed flawlessly), Imperfect (the cup is dropped onto the table rather than being placed carefully), Unaware (failed to pick the stir, and spun without a stir), Corrected (failed to pick the stir but went back to pick it after a few spinning), and Failure (the cup was dropped at the beginning). 
The results and further illustrations of five cases are shown in \autoref{online2}.
For scenarios other than Corrected, participants were asked to give 15 progress or scalar feedback. In Corrected, 20 progress or scalar feedback were collected.  
% Each point in \autoref{online2} represents the average preference or progress of 18 participants.
\subsection{Quantitative analysis}
To analyze the data, we used t-tests \cite{witte2017statistics} and Bayesian statistics with the schemes present in \cite{van2021jasp}.
For t-test results, 
we use Shapiro-Wilk tests to determine if the data is from a normal distribution. 
If the data is from a normal distribution, we use a standard independent samples t-test. 
Otherwise, we apply Kruskal-Wallis H Tests and Wilcoxon Rank-Sum Tests to our results.
For Bayesian statistics, 
a Bayes Factor (BF) is used. 
We interpret BF lower than 3 as ``no evidence''  for the alternative
hypothesis, between 3 to 10 as ``moderate evidence'', and 30 or above as ``strong evidence''.

\subsection{Giving progress is not time-consuming and not hard}
\begin{figure}
    \centering
    \includegraphics{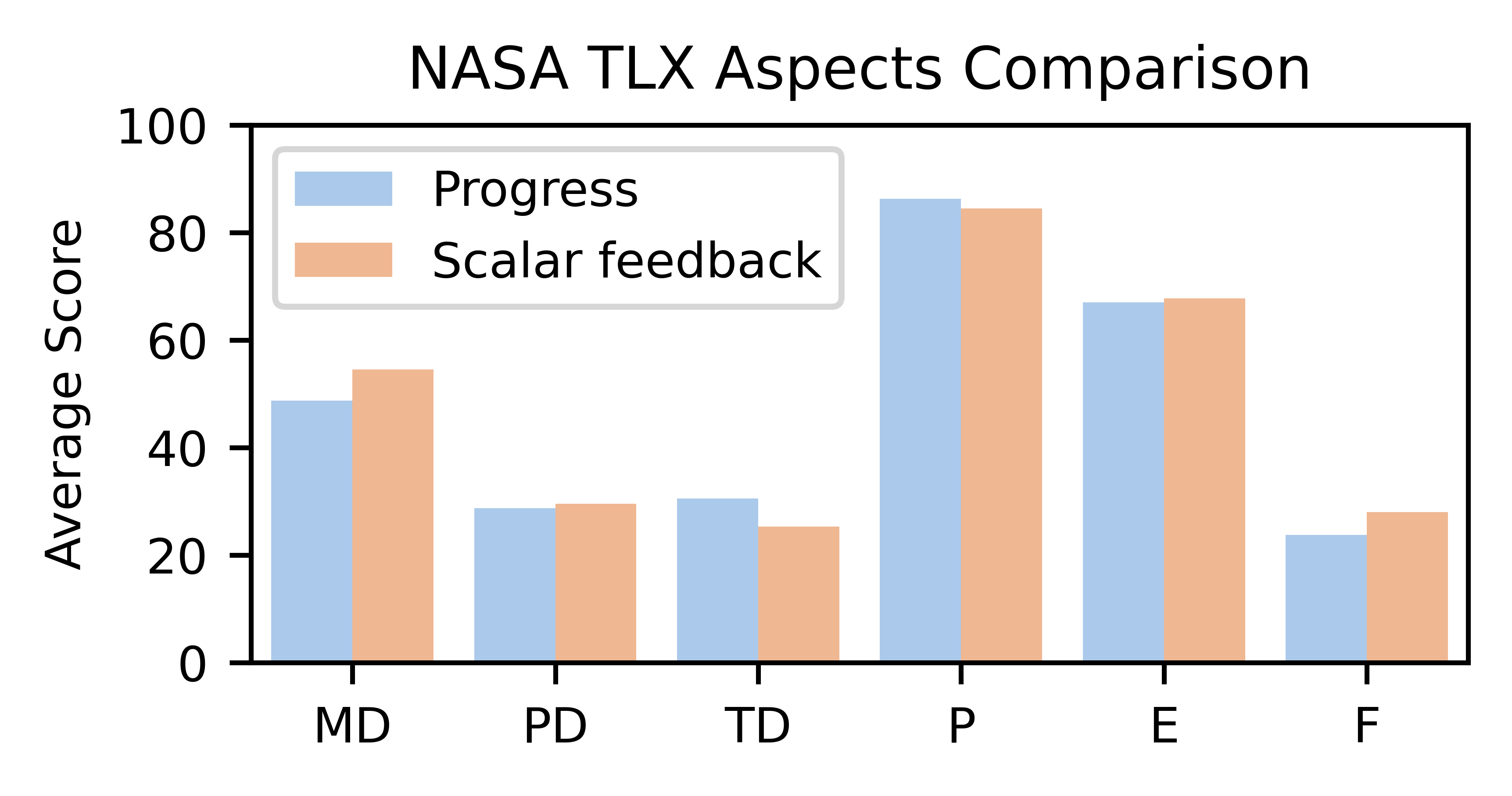}
    \caption{Online Study One Results. The workload of giving progress has no difference from giving scalar feedback.}
    \label{fig:phase1}
\end{figure}

We used the NASA TLX form to measure the workload, and we recorded the average task completion time.
The results are shown in \autoref{fig:phase1}. 
We did not find any significant difference between giving scalar feedback and progress in all the dimensions of the NASA TLX results using t-tests and BF ($p > 0.45$ and $0.3 < BF < 1$ for all dimensions). 
The average task completion time for evaluating the demonstrations using progress is 22 minutes 3 seconds, and the average task completion time for using scalar feedback is 23 minutes 10 seconds. The average completion time is lower for progress by $5.06\%$, but we did not find any significant difference using t-test and BF ($ p =0.442$, and $BF = 0.444$). \textbf{H1} is supported. 

\begin{figure*}
    \centering
    \includegraphics[width = 0.9\textwidth]{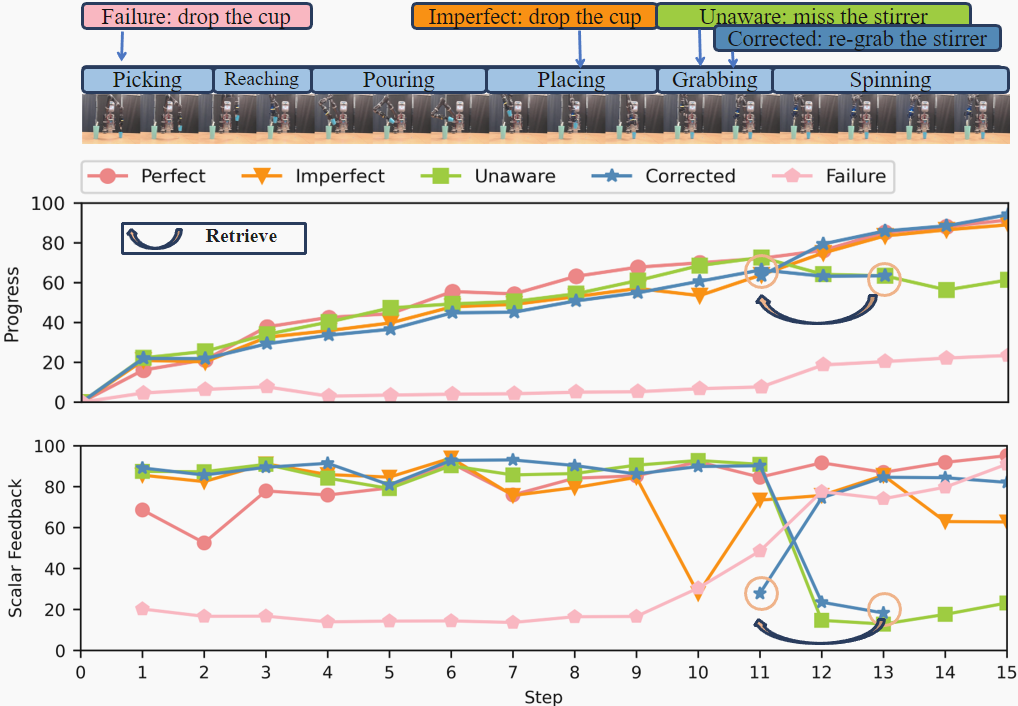}
    \caption{Online Study \RNum{2} Results.
    % Progress and scalar feedback are collected over five scenarios, Perfect, Imperfect, Unaware, Corrected, and Failure. 
    Progress and scalar feedback carry different information. 
    Scalar feedback reflects the optimality of a trajectory, while progress reflects the degree towards completing the task.
    Progress is more consistent than scalar feedback when the demonstration is non-optimal across participants.
    Progress is also capable of indicating if the task is completed successfully even if the robot has made a minor mistake or made a faulty mistake but fixed it. 
    }
    \label{online2}
\end{figure*}

% \subsection{Study Phase 2 Results: Progress  Indicates Task Completions}

% % \hang{one thing here: imperfection demonstration. progress tells you the quality of demonstrations. }
% % Study phase one indicates that progress is not only time and does not require extra effort from participants compared to preference. 
% % In this section, we ran a closer investigation of an integrated task with imperfect demonstrations to show the capability of scalar feedback and progress.
% In phase one, we showed that progress could indicate task completion with simple tasks and perfect demonstrations. 
% However, demonstrations, especially demonstrations from naive users, are not always perfect. 

% Thus, 

% #\subsection{Phase 2 Results: Progress indicated task completion}

% It is extremely hard for humans to provide perfect demonstrations to complex control tasks.
% It is possible that we ask a demonstrator to complete a series of tasks and the demonstrator fails one or two tasks. 
% In our error case and non recover case, the robot successfully finished part of the task but the task is still incomplete. 
% Preference indicates that the robot performed bad after step 12 in non recover case and before step 11 in error case, but 
% it is impossible to tell if the task is complete or not. 
% Progress, on the other hand, clearly indicates that the task is incomplete and the robot complete the task more in the non recover case. 

\subsection{
Progress is used to signify the completion rate of a task}
% We collected progress and scalar feedback from two groups of participants over five representative scenarios.  
The results of online study \RNum{2} are shown in \autoref{online2}. Each line represents the average progress or scalar feedback of 18 participants under a certain scenario. 
While scalar feedback and progress can both describe the quality of a demonstration, progress and scalar feedback focus on different information. 

Progress describes the degree to what extent a task has been completed. 
Progress was low at the beginning, increased as the task was completed more, stayed the same or became lower while the robot was not acting towards finishing the task (see Failure before spinning, Unaware after missing the stirrer, and Corrected before re-picking the stirrer).  A higher value of progress indicates more successful task completion. 
For instance, in the Unaware scenario, the progress at the final step is greater than in the Failure scenario but lower than in the Perfect scenario, due to 5 out of 6 sub-tasks being completed in the Unaware, as opposed to 1 out of 6 in the Failure.
%shown in \autoref{step15},$progress_{Pefect} \approx progress_{Imperfect} \approx progress_{Corrected} > progress_{Unaware} > progress_{Failure}$. 
We calculate Pearson corrections between the progress function from participants and an accumulative progress function created by true labels (+1 for completing part of the task, 0 for not completing anything, and -1 for backtracking) for each scenario. The correlations are 0.983 for Perfect scenario ($p< 0.001$), 0.982 for Imperfect scenario ($p< 0.001$), 0.974 for Unaware scenario ($p< 0.001$), 0.977 for Corrected scenario ($p< 0.001$), and 0.895 for Failure scenario ($p< 0.001$). 
\textbf{H2} is supported. 

Scalar feedback reflects the quality of a trajectory, and only a single trajectory. 
The average values of scalar feedback were generally high if there were no imperfections, and dramatically changed if there were any mistakes had been made, no matter if the mistakes would affect the robot completing the task (step 10, dropping cup, in Imperfect scenario) or if the robot did better in previous sub-tasks (step 12 to step 15 in Failure). 

\subsection{Progress indicates if a task is complete}
% \usepackage{multirow}
% \usepackage{multirow}
% \usepackage{multirow}
\begin{table}[]
\footnotesize
\begin{tabular}{ccccc}
\hline
                           &          & Average & IQR  &Comparison to Perfect                    \\ \hline
\multirow{2}{*}{Perfect}   & Progress & 91.5    & 2.75 &  \\
                           & Scalar F & 95.1    & 0.75 &                                                \\ \hline
\multirow{2}{*}{Imperfect} & Progress & 89.0    & 15.0 &   $p = 0.71$, $BF = 0.34$  \\
                           & Scalar F & 62.7    & 43.75 &    $p < 0.01$,$ BF >1000$                                        \\ \hline
\multirow{2}{*}{Unaware}   & Progress & 61.3    & 21.25 & $p < 0.01$, $BF = 98.75$\\
                           & Scalar F & 23.1    & 50.0 &    $p < 0.01$, $BF > 1000$                                      \\ \hline
\multirow{2}{*}{Corrected} & Progress & 94.1    & 8.25 &   $ p = 0.62$, $BF = 0.33$     \\
                           & Scalar F & 82.0    & 23.75 &   $p < 0.05$, $BF = 2.01$                                        \\ \hline
\multirow{2}{*}{Failure}   & Progress & 23.4    & 17.5 & $p < 0.01$, $BF > 1000$  \\
                           & Scalar F & 91.0    & 10.0 & $p = 0.30$, $BF = 0.46$                                      \\ \hline
\end{tabular}
\caption{Average progress and scalar feedback at the last step. Progress shows the ability to indicate task completion even if the demonstration is not perfect.}
\label{step15}
\end{table}

% \hang{combine }
We showed average progress and average scalar feedback at the last step for all participants and statistical analysis results between each scenario and the Perfect scenario in \autoref{step15}.
We found that progress could correctly indicate if the task is complete, while the indication of task completion was not captured by scalar feedback.  
% For Imperfect scenario and Correct scenario, the robot was not acting perfectly while the task was successfully completed. 
The task has been completed in three scenarios, Perfect, Imperfect, and Corrected. 
% The cup was dropped onto the table instead of carefully placed in the Imperfect scenario.
% The stirrer was not picked up initially and the robot has to re-attempt to pick after a few unproductive spinning.
The average progress of three completed scenarios at the last step is about 90, and 
there was no evidence showing that there is any difference between the other two and the Perfect case ($ p = 0.714, BF = 0.34$ for Imperfect, and $ p = 0.620, BF=0.33 $ for Corrected).
For the Unaware and Failure scenarios, there is strong evidence indicating that the task was not completed ($avg = 61.3, p<0.001, BF = 98.75$ for Unaware, and $avg = 23.4, p<0.001, BF>1000$ for Failure).
\textbf{H3} is supported.

% For the scenarios that the task has been successfully completed,
% the average progress is approximately 90, which we will use in our public space study to distinguish if a task has been completed. 
% In the Imperfect scenario, the cup was not carefully placed back on the table but that does not prevent the robot from completing the task, as a result there is no difference between progress at the last step with the Perfect scenario.  

% We showed progress and scalar feedback at the last step for all participants in \autoref{}.
% For the Perfect scenario, progress and scalar feedback are similar ($avg_{progress} = 91.5$, $avg_{scalar feedback} = 95.1$, $p-value =0.494$, $BF = 0.36$). 

\subsection{Progress is robust and more consistent to sub-optimality}
% Moreover,
% progress better describes unproductive but harmless behaviors. 
In the Imperfect scenario, when the robot dropped the cup onto the table, the progress remained at a similar level and did not affect the eventual progress.
The conclusion holds the same in the Corrected scenario. The progress only changed slightly when the robot missed the stirrer and started spinning unproductively, and progress at the end is similar to Perfect.
Scalar feedback, on the other hand, changed dramatically in all these cases. Participants used scalar feedback to indicate if a demonstration was clean and good, but the possibility that the errors were "harmless explorations" or "fixable mistakes" is not captured by scalar feedback. 
Moreover, for every scenario other than Perfect, progress has a lower average standard deviation in each scenario compared to scalar feedback. 
The average standard deviation for Imperfect: 16.1 for progress and 22.5 for scalar feedback ($ p = 0.018$, $BF = 3.19$), Unaware: 16.8 for progress and 21.1 for scalar feedback ($ p = 0.060$, $BF = 1.41$), Corrected 16.1 for progress and 20.3 for scalar feedback ($ p = 0.058$, $BF = 1.39$), and Failure:  15.6 for progress and 28.3 for scalar feedback ($p < 0.001$, $BF >1000$). \textbf{H4} is supported.

% \subsection{Phase 2 Results: Progress increased more if one sub-task is done}

% \subsection{Phase 2 Results: Progress is robust to imperfections during the task performing}

% While delta progress correlates preference ( \( r = \) ), the objectives of progress and preference are different. 
% In recover and non recover cases step 11 to 12, in which the robot missed the stirrer, the average preference significantly dropped from a high value to a low value, while the average progress stayed similar. 
% Similar phenomenon can also be observed when the robot dropped the cup instead of placing (step 10 in imperfect).
% Preference better describes whether the robot is performing correct behaviors by having dramatic changes in values, and progress loyally reports whether the robot's behaviors are beneficial towards the completion.  

% \subsubsection{Progress is robust to imperfections in demonstrations}
% In the imperfect demonstration, the robot dropped the cup to the table instead of placing the cup but dropping the cup did not effect the robot to finish the task.
% In the recover demonstration, the robot missed the stirrer for once and re-grabbed the stirrer after spinning without a stirrer. 
% In these two demonstrations, the robot made mistakes but the task was still fully completed.
% With preference, even the mistakes could be ignored or already recovered, the average preference values are lower (\hang{calculate it}) than the preference values before the mistakes happened, while the progress values in the end are the same to the progress value of the perfect demonstration. 

\section{Progress with Self-Provided Demonstrations from Non-experts}
While progress can be effective in being used along with expert demonstrations,
using progress to annotate non-expert demonstrations may still be challenging.
We expected most hypotheses will hold. 
% and progress will be more consistent across participants than scalar feedback if demonstrations are imperfect and multi-modal. 
We conducted a public space study to validate the applicability of progress with self-provided demonstrations from non-experts. 
% We have [XX] participants provided progress and preference over a robot candy scooping task. 
% Participants were from a university building and the overall participation time was about 15 minutes. 
% The study was approved by the Institution of Reviewer Board. 
We recruited 40 participants to provide demonstrations, progress, and scalar feedback in an ice cream topping-adding task.
Participants were recruited from the atrium of a university building and the overall participation time was about 15 minutes. 
This work is approved by the Institutional Review Board and all data collected was anonymous.  
We released all data we collected from our public space study as a dataset, along with the example scripts that read the data from files. The dataset is available at: \url{https://github.com/TeachingwithProgress/Non-Expert_Demonstrations}
% \hang{Not sure how we gonna do this}

\subsection{Experiment Setup}
The study was settled in the lobby of a university building. 
Each participant was asked to first give one demonstration and then watch a replay of the demonstration.
The task we asked participants to demonstrate is an ice cream topping-adding task. 
The goal for participants is to pick up a topping from a shelf and pour the topping into an ice cream via teleoperating a robot arm.  
The shelf is located on the right side of the workspace, and 
there are four toppings available which are located at four locations. 
The participants controlled the arm by using an Xbox controller. 
The arm was a Kinova Gen 3 Lite arm with six DoF. 
The setup and the workspace are shown in \autoref{page1}.
We recorded the demonstrations in 5 HZ. 
During the replay, the arm would stop every 10\% of the frames, and we would ask participants for one progress and one scalar feedback for the replayed partial trajectory. 
% The recorded frames contained arm's joint positions, gripper positions, object positions, and the target zone position. 

% Before giving a demonstration, participants could practice the task for three minutes.  
% While they were giving demonstrations, 

% Once they started giving demonstrations, they were not able to   

% After recording their demonstration, we asked them to watch a replay of their demonstration.  
% During the replay, the arm would stop after replaying 100 frames (one sub-trajectory) and ask participants for progress, and scalar feedback. 

\subsection{Experiment Procedure}
We recruited participants by asking people who walked by our setup. 
Of the 40 participants, 22 participants were male, 14 participants were female, and 4 participants preferred not to say. 
If the participants agreed to join the study,
we first asked them if they were familiar with robots, and excluded them if they said yes. We
then asked them  
to fill out a consent form.
Then we introduced them to the ice cream topping adding task, and how to use an Xbox controller to teleoperate the arm. 
Participants had up to 3 minutes to practice the task before giving a demonstration. 
Each participant only had one chance to give a demonstration and could not retry.
After giving a demonstration, the experimenter introduced participants to the replay evaluation task.
The experimenter would introduce progress and scalar feedback in detail to reduce the difference in understanding of the signals among participants. 
During the replay, the arm would stop ten times.
For each time, the experimenter would briefly explain progress and scalar feedback again, and ask participants for two signals in a random order. 

% Then, the replay started and participants evaluated replays via keyboard by typing the numbers.
% In the end, participants were asked to fill another questionnaire, end of the study. 

% One key takeaway from our online study is that demonstrations are not necessarily provided by users and 
% users were capable to appropriately address whether the task was correctly ongoing or finished with other's demonstrations. 
% since progress provides insights about the quality of a demonstration and the weight of each subtrajectory, we believe that 
% Moreover, we believe that users should be able to provide better progress to their own demonstrations. 

\subsection{Non-Expert Demonstrations Are Multi-Policy and Noisy}
\label{sectionVC}
\begin{figure}
    \centering
    \includegraphics[width = 0.45\textwidth]{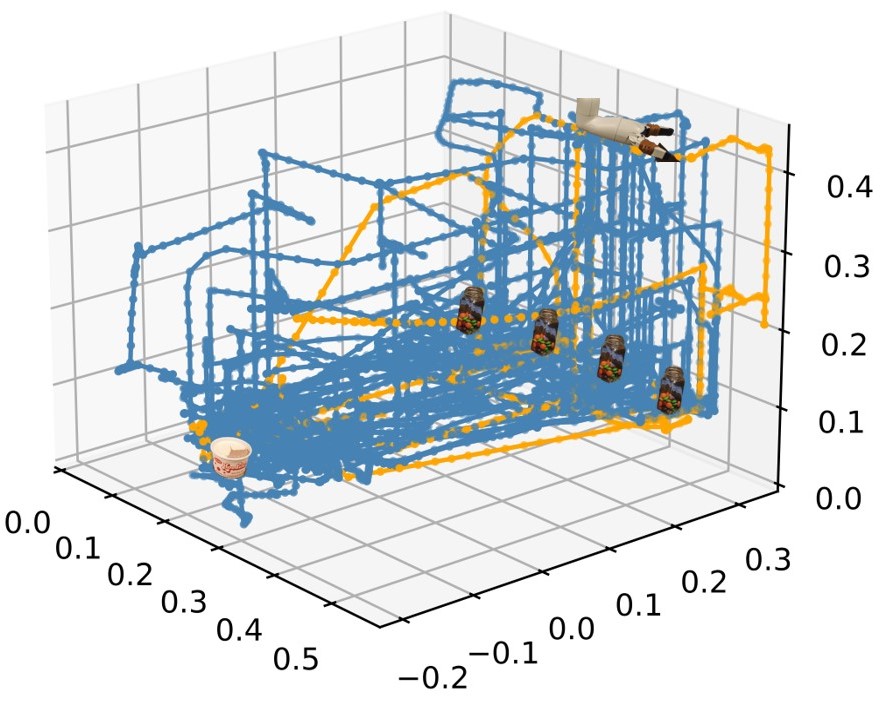}
    \caption{3D visualization of 40 non-expert demonstration trajectories. The positions of objects are marked out with images. The blue trajectories are successful demonstrations and the orange trajectories are faulty demonstrations. Most of the demonstrations succeed, while the policies are diverse.}
    \label{3d traj}
\end{figure}
% In prior work, 
% We collected 

We collected 40 demonstrations from 40 participants. All trajectories are visualized in \autoref{3d traj}.
We also visualized the locations of the objects used in the experiment. 
The blue trajectories are successful demonstrations and the orange trajectories are failed demonstrations. 
We found that non-expert demonstrations are noisy and contain a variety of policies, even though 34 out of 40 are ultimately successful. 
This suggests that policies can be both successful and sub-optimal, which supports our intuition: assessing the quality of demonstrations by comparing is likely to lose information if there are many "good enough" policies. 
We also observed that the noise in the demonstrations is not only teleoperating errors but also explorations.
For instance, we observed that 14 participants slightly shook the topping jar to test if the gripper firmly held the jar when picking the jar, and 17 participants poured a few toppings out first to see if the jar was right above the ice cream when pouring the toppings. 
This highlights the importance of detecting unproductive behaviors,
and suggests that \textbf{noisy demonstrations from partially trained agents or perfect demonstrations with injected noise are inappropriate approximations of noisy human demonstrations}: human demonstrations are ``noisy'' in a meaningful way. 
For the six failed demonstrations, 
three of them were because the topping jar was accidentally dropped while reaching the ice cream, and two of them were because the gripper did not successfully pick up the topping jar.
The most common faulty cases are similar to the Failure scenario and the Unaware scenario we used in our online study, which confirmed the validity of the design of our online study.

\subsection{Progress indicates task completions and is more consistent across participants than scalar feedback }
\begin{figure}
    \centering
    \includegraphics{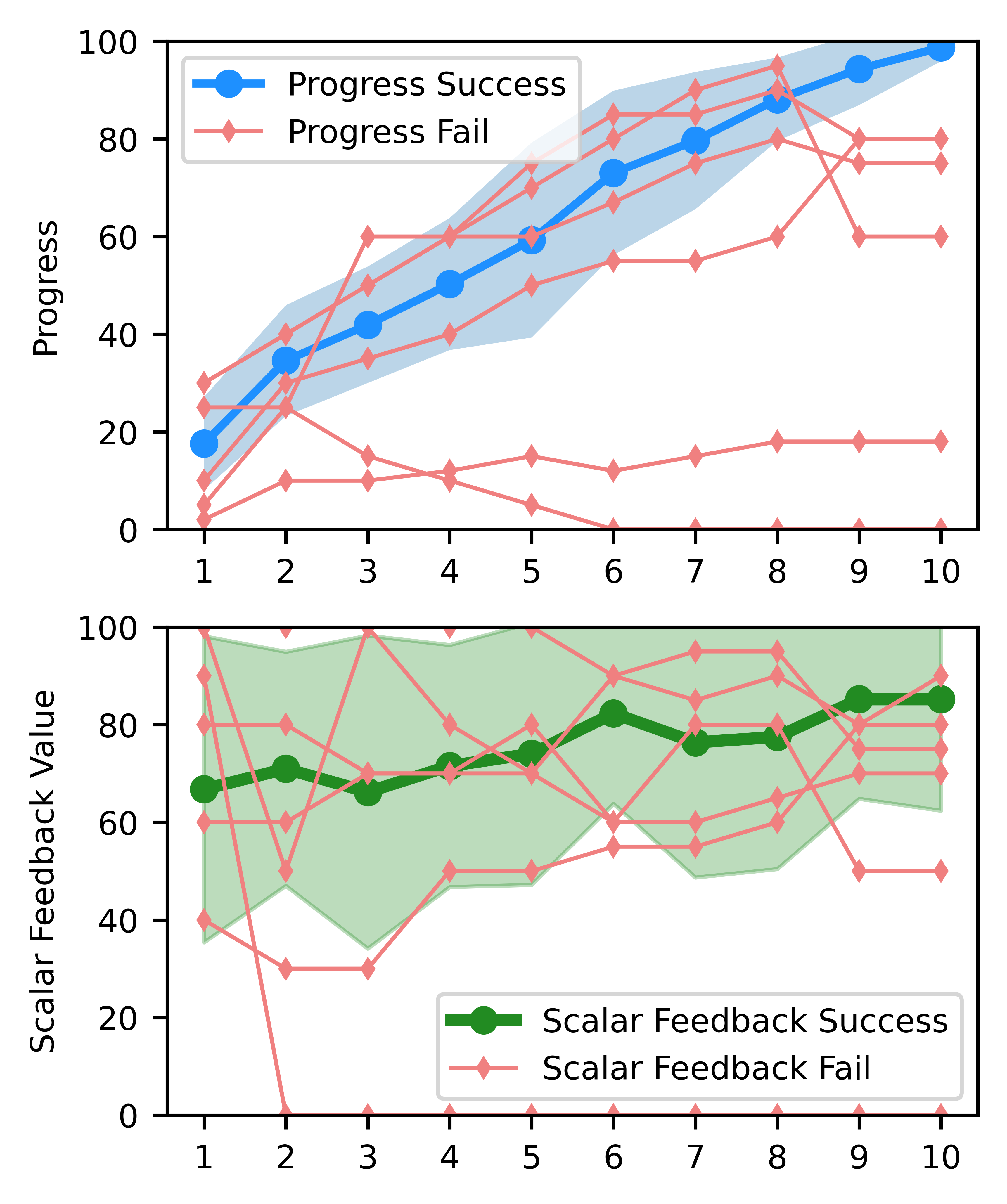}
    \caption{Progress and scalar feedback over 32 Demonstrations. Progress increases as the task approaches completion, indicates task success, and is consistent among participants.  }
    \label{fig:suc_fail}
\end{figure}

We collected progress and scalar feedback from 34 participants (two participants' progress and scalar feedback were excluded since all progress and scalar feedback they provided were 100).
The results are shown in \autoref{fig:suc_fail}. 
We plot the average progress and scalar feedback for all successful demonstrations, and progress and scalar individually for all failure demonstrations.
We are not able to determine if progress from participants correctly describes the task completion rate since we do not have ground truth labels, but average progress for successful demonstrations did start from low and increased as the demonstrations were reproduced which is a strong indication that progress correlates with task completion rates.

We successfully identified all failed demonstrations by only looking at progress at the last step. 
We use 90 as the divide value, which is the average progress at the last step for successful demonstrations in our online study. 
If progress at the last step is less than 90, the task is incomplete, otherwise the task is completed. 
For all failed demonstrations, the end progress is less than 90 and the average progress at the end for successful demonstrations is 98.8.  \textbf{H3} is supported. 
Moreover, progress is significantly more consistent across participants even though their demonstrations are multiple policies and in different quality ($avg\_std_{progress} = 11.6$, $avg\_std_{scalar feedback} = 25.5$, $p < 0.001$, $BF >1000$).
\textbf{H4} is supported.

% \subsection{Progress Well Aligned with Learned Reward Function}

\subsection{Progress allows the awareness of backtracking}
\begin{figure}
    \centering
    \includegraphics[width = 0.4\textwidth]{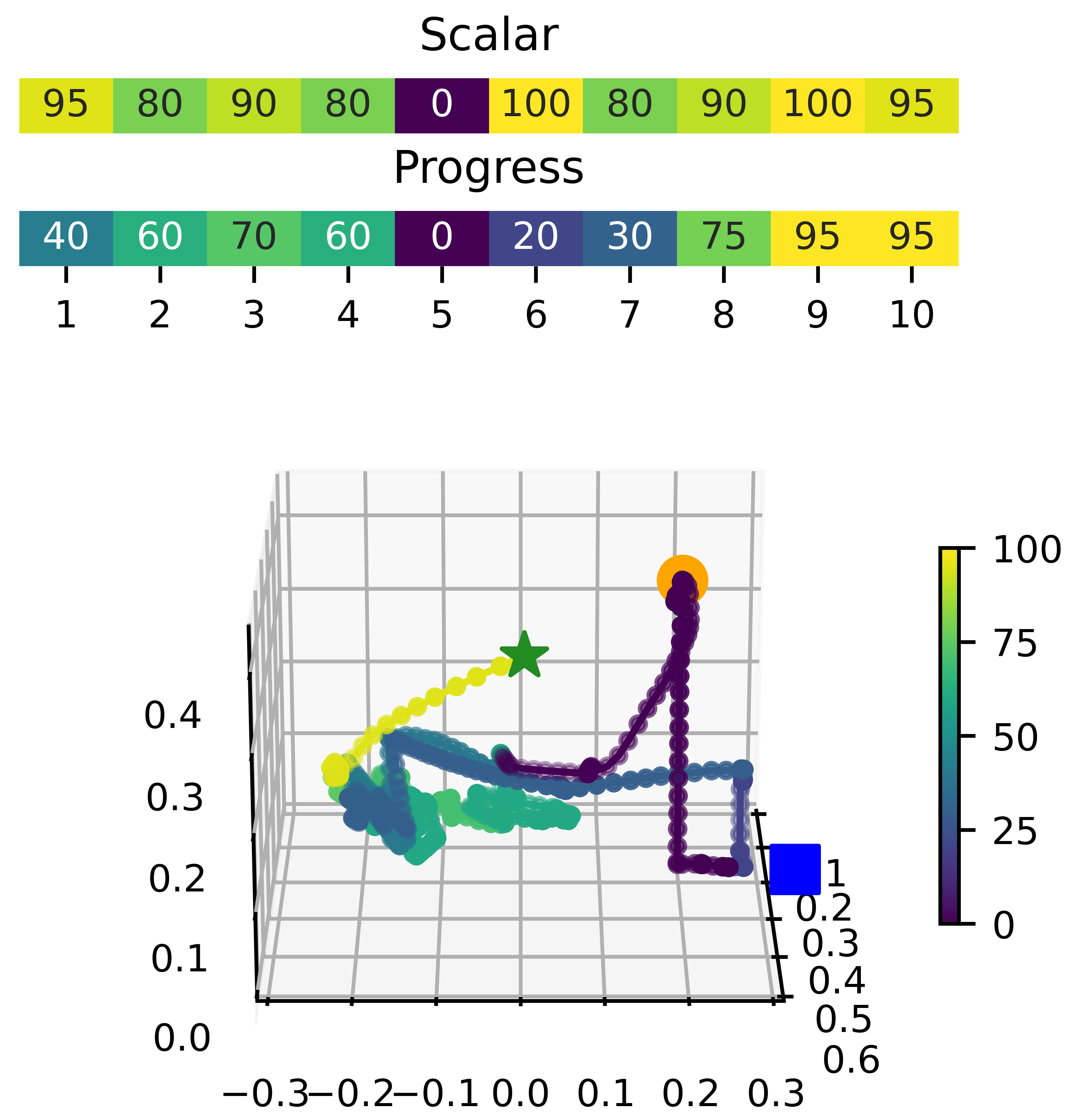}
    \caption{3D trajectory of one participant's demonstration. The trajectories are colored with progress, and both scalar feedback and progress signals are indicated in a timeline above the figure. The brown circle represents the start point. The green star represents the endpoint. The blue square is the position of the topping jar. The participant reset the arm to the start position after demonstrating a few steps to adjust the grip point.}
    \label{fig:p31}
\end{figure}
As mentioned in \autoref{sectionVC}, participants would do explorations while demonstrating difficult parts of the task.
We also observed that some participants went back a few steps to adjust the grip pose or the pouring position.
For instance, one participant failed to pour the toppings into the ice cream because the grasp point was not optimal. 
Then the participant decided to reset the arm to the initial position, demonstrated the entire task from the beginning again, and succeeded.
We plot the trajectory for that demonstration along with progress and scalar feedback, shown in \autoref{fig:p31}.
As indicated by progress, the arm was reset to the initial position between step 4 and step 5, and the task was successfully completed afterward.

\begin{figure}[t]
    \centering
    \includegraphics{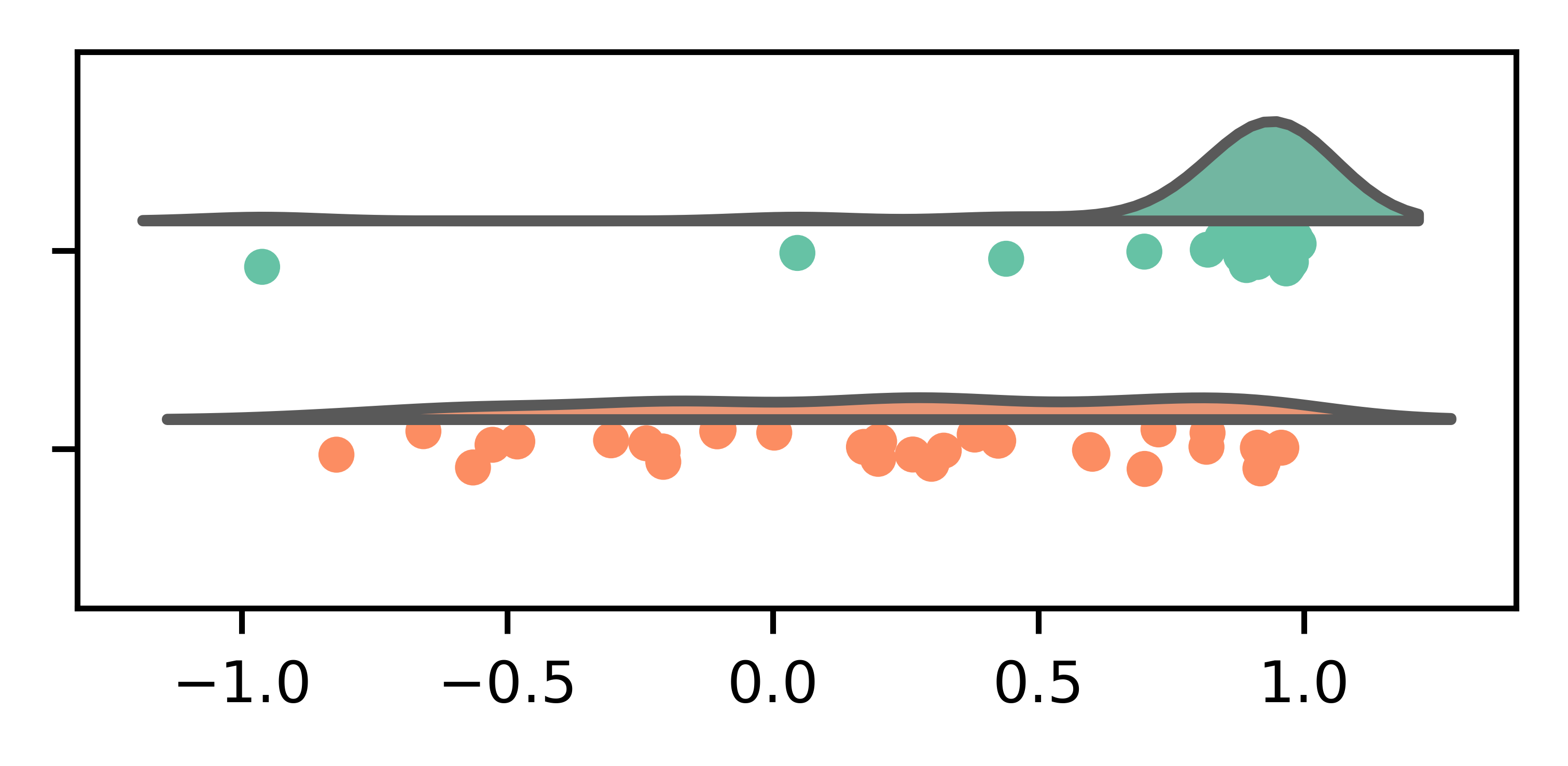}
    \caption{Correlations between progress and rewards (top), and correlation between scalar feedback and rewards (bottom) over 32 participants. The correlations between progress and rewards have an average of 0.83.}
    \label{fig:cor-pro-sca}
\end{figure}

\section{Discussion}
In this work, we investigated and closely defined an under-explored teaching signal, \emph{progress}, and conducted three different studies to show the usefulness of progress across a variety of scenarios. 
We showed that giving progress is not hard and progress carries information beyond scalar feedback and other teaching signals in prior work. 
We expect that progress will be effective in many applications other than just using along with demonstrations. 

\textbf{Prevent Reward Hacking}
Progress could be a powerful signal to indicate reward hacking. Reward hacking is a phenomenon in which a learning agent learns to achieve high rewards by performing unintended actions instead of finishing the task. For instance, a cleaning robot gets a +1 reward every time it cleans a room. The robot, instead of cleaning one room and going to the next room, repetitively ejects dirt in a room, then cleans it, and thus achieves high rewards. 
Using progress, we can easily identify that the robot is not advancing towards task completion. 

\textbf{Inverse Reinforcement Learning}
We observed a similarity between progress functions and reward functions and expect that a progress function might be a representation of a reward function. 
We trained a reward function using demonstrations we collected in the public space study using Adversarial Inverse Reinforcement Learning \cite{fu2018learning}, and calculated the rewards for each demonstration. 
We then calculated the Pearson correlation between the rewards of the demonstration and the progress, as well as the correlation between the rewards and the scalar feedback from each participant in \autoref{fig:cor-pro-sca}.
We found that progress is strongly correlated with the learned reward function ($avg_{r} = 0.83 $) and the average correlation is significantly higher ($p < 0.001$, $BF > 1000$ ) than scalar feedback ($avg_{r} = 0.19$).

% \textbf{Weighted Behavior Cloning}
% The tasks we picked for our studies consist of multi-subtasks
% We noticed that increase of progress is higher when a sub-task

% \textbf{Mistake Recover}

\textbf{Data Filtering and Ranking}
Progress could also be used to rank demonstrations. 
For example, a demonstration with a progress of 60 at the end should be ranked lower than a demonstration with a progress of 90. 
Moreover, progress can be applied as a data filter especially when the demonstrations are sub-optimal. 
% For example, if a participant made a mistake in the middle of a demonstration and then fixed it, we could only filter out the part with decreasing progress labels, and keep the rest since we know the task is completed.  
% With scalar feedback, we would only know there was a mistake by using scalar feedback. 
% We failed to train a working agent using the behavior cloning from noisy demonstrations algorithm from  \citeauthor{sasaki2020behavioral}, but we did observe that the agent with data filtering by progress performed better.

A key area for future work is to compare the model performance between the model trained using progress and the model trained using other types of human feedback.  Training a reliable model from limited non-expert demonstrations and non-expert annotations is challenging but future work could expand our data with more demonstrations along with more types of human teaching signals such as preference. 

% \textbf{Guiding Exploration}

\section{Conclusion}
In conclusion, we defined \textit{progress} in detail and found that \textit{progress} could be used to describe completion degrees of a task, indicate if a task is complete, and be more consistent across users, without requiring extra workload or time compared to giving scalar feedback. 
We collected 40 non-expert demonstrations along with progress and scalar feedback, and released them as a dataset. 
We found that non-expert demonstrations are multi-policy and mostly successful, while noisy in a meaningful way. 
Our work suggests 
that progress is 
information-rich and is worth more attention to develop new methods to effectively leverage the novel information from progress. 

\section*{ACKNOWLEDGMENT}
The work described here was supported in part by the US National Science Foundation (IIS-2132887).

\bibliographystyle{IEEEtran}
\bibliography{references}

\end{document}